\documentclass{article}
\usepackage{lineno,hyperref}
\usepackage{amssymb}
\usepackage{multirow}
\usepackage{algorithm}
\usepackage{algorithmic}
\usepackage{color}
\usepackage{graphicx}
\usepackage{natbib}

\begin{document}

\title{Multi-lingual and Cross-lingual TimeLine Extraction\thanks{Preprint submitted to \emph{Knowledge Based Systems} 17 January, 2017.}}
\author{Egoitz Laparra\thanks{Corresponding author: \texttt{egoitz.laparra@ehu.eus}}, Rodrigo Agerri, Itziar Aldabe, German Rigau\\IXA NLP group\\University of the Basque Country (UPV/EHU)\\Donostia-San Sebasti\'an, Basque Country}
\date{}
\maketitle

\begin{abstract}
In this paper we present an approach to extract ordered timelines of events, their participants, locations and times from a set of multilingual and cross-lingual data sources. Based on the assumption that event-related information can be recovered from different documents written in different languages, we extend the Cross-document Event Ordering task presented at SemEval 2015 by specifying two new tasks for, respectively, Multilingual and Cross-lingual Timeline Extraction. We then develop three deterministic algorithms for timeline extraction based on two main ideas. First, we address implicit temporal relations at document level since explicit time-anchors are too scarce to build a wide coverage timeline extraction system. Second, we leverage several multilingual resources to obtain a single, interoperable, semantic representation of events across documents and across languages. The result is a highly competitive system that strongly outperforms the current state-of-the-art. Nonetheless, further analysis of the results reveals that linking the event mentions with their target entities and time-anchors remains a difficult challenge. The systems, resources and scorers are freely available to facilitate its use and guarantee the reproducibility of results.
\end{abstract}

\noindent \textbf{Keywords:} Timeline extraction, Event ordering, Temporal processing, Cross-document event coreference, Predicate Matrix, Natural Language Processing


\section{Introduction}\label{sec:introduction}

Nowadays, Natural Language Processing (NLP) may help professionals to access high quality, structured knowledge extracted from large amounts of unstructured, noisy, and multilingual textual sources \citep{VossenKBS2016}. As the knowledge required usually amounts to reconstructing a chain of previous events, building timelines constitutes an efficient and convenient manner of structuring the extracted knowledge. However, yielding timelines is a high level task that involves information extraction at multiple tiers, including named entities, events or time expressions. Furthermore, it should also be considered that the information required to construct the timeline must be gathered from different parts of a document, or even from different documents. Thus, coreferential mentions of entities and events must be properly identified.

For example, a named entity can be mentioned using a great variety of surface forms (Barack Obama, President Obama, Mr. Obama, Obama, etc.) and the same surface form can refer to a variety of named entities \footnote{For example, see \url{http://en.wikipedia.org/wiki/Europe_(disambiguation)}}. Furthermore, it is possible to refer to a named entity by means of anaphoric pronouns and co-referent nominal expressions such as `he', `her', `their', `I', `the 35 year old', etc. The same applies to event mentions, which can be verbal predicates or verbal nominalizations. Thus, the following two sentences contain different mentions of the same event, namely, that a gas pipe exploded, via the two different predicates `exploded' and `blast'. Furthermore, while Example (1) allows us to explicitly time-anchor the event via the temporal expression `yesterday', that does not occur in the second example. In this context, building a timeline amounts to detecting and temporal ordering and anchoring the events in which a target named entity participates.

\begin{itemize}
\item[(1)] A leak was the apparent cause of yesterday's gas \emph{blast} in central London.
\item[(2)] A gas pipe accidentally \emph{exploded} in central London. Only material damage was reported.
\end{itemize}

Temporal relation extraction has been the topic of different SemEval tasks \citep{VerhagenSemEval07,VerhagenSemEval10,UzZamanSemEval13,LlorensSemEval15} and other challenges as the 6th i2b2 NLP Challenge \citep{Sun2013}. These tasks have focused mainly on the temporal relations of events with respect to other events or time expressions, and their goal is to discover which of them occur before, after or simultaneously with respect to others.

Recently, SemEval 2015 included a novel task regarding temporal information extraction \citep{minard2015}. The aim of SemEval 2015 task 4 was to order in a timeline the events in which a target entity is involved. The task presents some significant differences with respect to previous evaluation settings. First, temporal information must be recovered from different sources across documents. Second, timelines are focused on the events pertaining just to a single given target entity. Finally, unlike previous challenges, SemEval 2015 task 4 requires a complete time anchoring.

In this work we build on the SemEval 2015 Timeline extraction task to present a system and framework to perform Multilingual and Cross-lingual Timeline Extraction. This is based on the assumption that timelines and events can be recovered from a variety of data sources across documents and across languages. In doing so, this paper presents a number of novel contributions.

\paragraph{Contributions} The original Cross-document event ordering task defined for SemEval 2015 (main Track A) is extended to present two novel tasks for two languages (English and Spanish) on Multilingual and Cross-lingual timeline extraction, respectively. The tasks also generated publicly available annotated datasets for trial and evaluation. Additionally, two new evaluation metrics improve the evaluation methodology of the SemEval 2015 task to address both the multilingual and cross-lingual settings.

Interestingly, we also show that the temporal relations that explicitly connect events and time expressions are not enough to obtain a full time-anchoring annotation and, consequently, produce incomplete timelines. We propose that for a complete time-anchoring the temporal analysis must be performed at a document level in order to discover implicit temporal relations. Furthermore, we show how to effectively leverage multilingual resources such as the PredicateMatrix \citep{predicatematrix} and DBpedia\footnote{\url{http://wiki.dbpedia.org/}.} to improve the performance in a more realistic setting of building cross-lingual timelines when no parallel data as input is available. We present a deterministic approach that obtains, by far, the best results on the main Track A of SemEval 2015 task 4. Our deterministic approach is fledged out via three different timeline extraction systems which extend an initial version presented in \cite{LaparraACL2015}. To guarantee reproducibility of results we also make publicly available the systems, datasets and scripts used to perform the evaluations \footnote{\url{http://adimen.si.ehu.es/web/CrossTimeLines}}.

Next section reviews related work, focusing on the SemEval 2015 Timeline extraction task. Next, Section \ref{sec:mult-crossl-timel} describes the two new Cross-lingual and Multilingual Timeline extraction tasks. The construction of the datasets for the new tasks occupies Section \ref{sec:annotation} and Section \ref{sec:scorer} formulates the evaluation methodology employed in this work. In section \ref{sec:evaluation} we report the evaluation results obtained by the systems previously presented in Section \ref{sec:system}. Finally, Section \ref{sec:error} provides an error analysis to discuss the results and contributions of our approach while Section \ref{sec:concluding-remarks} highlights the main aspects of our work and future directions.

\section{Related work}\label{sec:related}

The present work is directly related to the SemEval 2015 task 4, Timeline: Cross-document event ordering \citep{minard2015}. Its aim is to combine temporal processing and event coreference resolution to extract from a collection of documents a set of timelines of events pertaining to a specific target entity. The notion of event is based on the TimeML definition, namely, an event is considered to be a term that describes a situation or a state or circumstance that can be held as true \citep{PustejovskyIWCS03}.

In fact, the Timeline extraction task is in turn quite close to the TempEval campaigns \citep{VerhagenSemEval07,VerhagenSemEval10,UzZamanSemEval13,LlorensSemEval15}. Briefly, the problem is formulated as a classification task to decide the type of temporal link that connects two different events or an event and a temporal expression. For that reason, the task has been mainly addressed using supervised techniques. For example, \cite{ManiACL06,ManiTechRep07} trained a MaxEnt classifier using training data which was bootstrapped by applying temporal closure. \cite{ChambersACL07} focused on event-event relations using previously learned event attributes. More recently, \cite{DSouzaNACL13} combined hand-coded rules with some semantic and discourse features. \cite{LaokulratSemEval13} obtained the best results in TempEval 2013 annotating sentences with predicate-role structures, while \cite{MirzaEACL14} state that using a simple feature set results in better performances. Other recent works such as \cite{ChambersTACL14} have pointed out that these tasks cover just a part of all the temporal relations that can be inferred from the documents.

The SemEval 2015 timeline extraction task proposed two tracks, depending on the type of data used as input. The main track A for which only raw text sources were provided, and Track B, where gold event mentions were also annotated. For each of the two tracks a sub-track was also proposed in which the assignment of time anchoring was not taken into account for the evaluation. No training data was provided for any of the tracks.

Track A received three runs from two participants: the WHUNLP and SPINOZAVU teams. Both approaches were based on applying a pipeline of linguistic processors including Named Entity Recognition, Event and Nominal Coreference Resolution, Named Entity Disambiguation, and temporal processing \citep{minard2015}. The SPINOZAVU system was further developed in \cite{caselli2015}.

The Track B approaches, represented by the two participants HEIDELTOUL and GPLSIUA, substantially differ from those of Track A because the event mentions pertaining to the target entity are already provided as gold annotations. Therefore, those systems focused on event coreference resolution and temporal processing \citep{minard2015}. Two recent works have been recently published on Track B: an extension of the GPLSIUA system \citep{NavarroColorado2016244}, and a distant supervision approach using joint inference \citep{Cornegruta2016}.

Track A is, in our opinion, the most realistic scenario as systems are provided a collection of raw text documents and their task is to extract the timeline of events for each of the target entities. More specifically, the input provided is a set of documents and a set of target entities (organization, people, product or financial entity) while the output should consist of one timeline (events, time anchors and event order) for each target entity.

Compared to previous works on Track A of the SemEval 2015 Timeline extraction task, our approach differs in several important ways. Firstly, it addresses the extraction of implicit information to provide a better time-anchoring \citep{PalmerACL86,WhittemoreACL91,TetreaultISRRNLP02}. More specifically, we are inspired by recent works on Implicit Semantic Role Labelling (ISRL) \citep{GerberCL12} and, specially, on \cite{BlancoEACL14} who adapted ISRL to focus on modifiers, including temporal arguments, instead of core arguments or roles. Given that not training data is provided, we developed a deterministic algorithm for timeline extraction loosely inspired by \cite{LaparraACL13}. Secondly, we extend the monolingual approach to make it multi- and cross-lingual, which constitutes a novel system on its own. Finally, our approach outperforms every other previous approach on the task, almost doubling the score of the next best system.

\section{Multilingual and Cross-lingual Timeline Extraction}\label{sec:mult-crossl-timel}

\begin{figure*}[ht!]
  \centering
  \includegraphics[width=120mm]{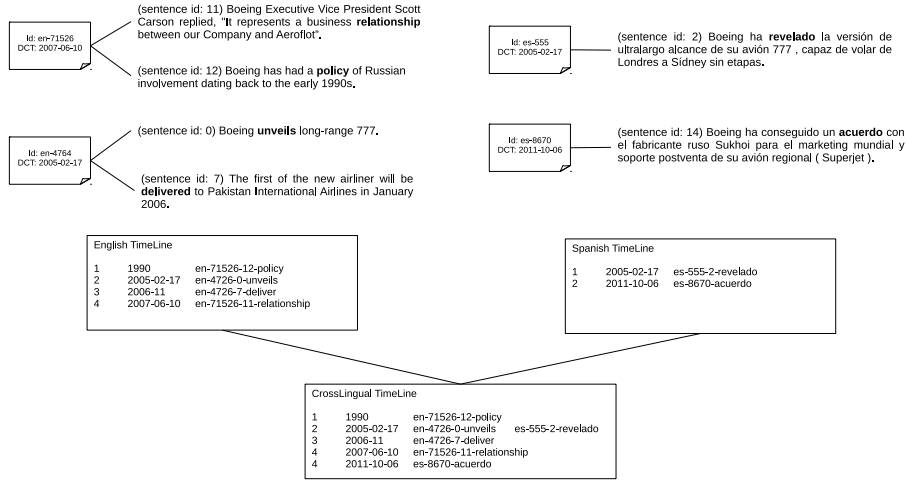}
  \caption{\label{fig:timeline} Example of multilingual and cross-lingual timelines for the target entity {\it Boeing}.}
\end{figure*}

The Timeline Extraction definition was formulated as follows: ``Given a set of documents and a target entity, the task is to build an event timeline related  to that entity, i.e. to detect, anchor in time and order the events involving the target entity'' \citep{minard2015}. As we have already mentioned in the previous section, in this work we will focus on Track A (main track), which is the most demanding and realistic setting of the two: systems are given a set of raw text documents and the task is to extract the timelines. Furthermore, we contribute two novel extensions to the original task:

\begin{itemize}
\item \textbf{Multilingual Timeline Extraction}: This task straightforwardly extends the SemEval 2015 task to cover new languages. Thus, a parallel set of documents and a set of target entities, common to all languages, are provided. The goal is to obtain a timeline for each target entity in each language independently.
\item \textbf {Cross-lingual Timeline Extraction}: For this task, the timelines are built from source data in different languages identifying those event mentions that are coreferent across languages. However, unlike in the multilingual setting, every document in every language is considered together so that \emph{a single cross-lingual timeline} is expected for each of the target entities.
\end{itemize}

These two new tasks are presented here for two languages, namely, English and Spanish. Figure \ref{fig:timeline} shows an example of both multilingual and cross-lingual timelines for the target entity \textit{Boeing}. The left-hand side column corresponds to an English timeline extracted from four sentences in two different English documents. On the right-hand side is shown an Spanish timeline obtained from two sentences contained in two different documents. Words in bold refer to the event mentions that compose the timeline. Finally, the box in the bottom depicts a cross-lingual timeline built from sources in both English and Spanish. Coreferent events across languages, such as {\it unveils} and {\it revelado}, are annotated in the same row, while events that are simultaneous but are no coreferent appear in different rows. The events {\it relationship} and {\it acuerdo} (in the last two rows) provide such an example. The following section describes in more detail the procedure used to build the datasets for both the Multilingual and Cross-lingual Timeline Extraction tasks.

\section{Data Annotation}\label{sec:annotation}

In the original Timeline Extraction task at SemEval 2015 \citep{minard2015}, the dataset was extracted from the raw text of the English side of the MeanTime corpus \citep{MEANTIME:2016}. Given that MeanTime is a parallel corpus that includes manual translations from English to Spanish, Italian and Dutch, it is straightforward to use its Spanish part for the Multilingual and Cross-lingual Timeline Extraction tasks.

\subsection{Creation of multilingual and cross-lingual timelines}\label{sec:creation}

In order to better understand the procedure to create the datasets for the multilingual and cross-lingual settings, a brief overview of the original annotation to create the gold standard timelines for English is provided. For full details of the original annotation, please check the SemEval 2015 task description \citep{minard2015}. As already mentioned, the input to the task consisted of the target entities, the event mentions and the time anchors. In the following, each of these three aspects are described.

\paragraph{Target Entities} A set of target entities were selected that belong to type PERSON (\textit{e.g. Steve Jobs}), ORGANISATION (\textit{e.g. Apple Inc.}), PRODUCT (\textit{e.g. Airbus A380}), and FINANCIAL (\textit{e.g. Nasdaq}). The target entities must appear in at least two different documents and be involved in more than two events.

\paragraph{Events} The annotation of events was restricted by limiting the annotation to events that could be placed on a timeline. Adjectival events, cognitive events, counter-factual events, uncertain events and grammatical events were not annotated. Furthermore, timelines only contain events in which target entities explicitly participate as \textit{Agent} or \textit{Patient}.

\paragraph{Time anchors} A time anchor corresponds to a TIMEX3 of type DATE \citep{PustejovskyCL03}. Its format follows the ISO-8601 standard: YYYY-MM-DD (i.e. Year, Month, and Day). The finest granularity for time anchor values is DAY; other granularities admitted are MONTH and YEAR (references to months are specified as YYYY-MM and references to years are expressed as YYYY).

\paragraph{Creation of multilingual timelines} The process described above was followed to create timelines in Spanish. In both cases, English and Spanish, timelines are represented in tabulated format. Each row contains one event representing an instance of an event occurring at a specific time. The first column of each row indicates the position of the event in the timeline. The second column specifies the time-anchor of the event. Additional columns in the row, if any, refer to the different mentions of that event in the dataset. Each event mention is identified with the document identifier, the sentence number and the textual extent of the mention. The document identifier is in turn composed of a prefix specifying the language in which the document is written and its numerical identifier. If two events have the same time-anchor but they are not coreferent, they are placed on different rows. An example of multilingual annotations for English and Spanish is provided by Figure \ref{fig:timeline}.

\paragraph{Creation of cross-lingual timelines} We automatically cross the annotations from the English and Spanish parallel corpora. The resulting timelines have the same format as the original ones. More specifically, when two mentions of the same event in two different languages refer to the same event then they are included in the same row. The automatic mapping of annotations to construct the cross-lingual timelines was manually revised. A brief example of a cross-lingual dataset is illustrated by the box at the bottom of Figure \ref{fig:timeline}.

\subsection{Task dataset}\label{sec:datasets}

\begin{table*}[!h]
\centering
\begin{tabular}{c|l||r||rrr|r}
  \multicolumn{2}{c||}{} & \multicolumn{1}{c||}{Trial} & \multicolumn{4}{c}{Test}  \\
  \multicolumn{2}{c||}{} &  Apple Inc. & Airbus & GM  & Stock & Total  \\
\hline
\hline
  \multirow{10}{*}{\rotatebox{90}{English (SemEval-2015)}} & \# documents & 30 & 30 & 30  & 30 & 90 \\
  & \# sentences & 463 & 446 & 430 & 459 & 1,335 \\
  & \# tokens & 10,343 & 9,909 & 10,058 & 9,916 & 29,893 \\
  \cline{2-7}
  & \# event mentions & 178 & 268 & 213 & 276 & 757 \\
  & \# event instances & 165 & 181 & 173 & 231 & 585 \\
  & \# target entities & 6 & 13 & 12 & 13 & 38 \\
  & \# timelines & 6 & 13 & 11 & 13 & 37 \\
  & \# event mentions / timeline & 29.7 & 20.6 & 19.4 & 21.2 & 20.5 \\
  & \# event instances / timeline & 27.5 & 13.9 & 15.7 & 17.8 & 15.8 \\
  & \# docs / timeline & 5.7 & 5.2 & 4.1 & 9.1 & 6.2 \\
\hline
\hline
  \multirow{10}{*}{\rotatebox{90}{Spanish}} & \# documents & 30 & 30 & 30 & 30 & 90 \\
  & \# sentences & 454 & 445 & 431 & 467 & 1,343 \\
  & \# tokens & 10,865 & 10,989 & 11,058 & 11,341 & 33,388 \\
  \cline{2-7}
  & \# event mentions & 187 & 222 & 195 & 244 &  661 \\
  & \# event instances & 149 & 163 & 147 & 212 & 522 \\
  & \# target entities & 6 & 13 & 12 & 13 & 38 \\
  & \# timelines & 6 & 13 & 11 & 13 & 37 \\
  & \# event mentions / timeline & 31.2 & 17.1 & 17.7 & 18.8 & 17.9 \\
  & \# event instances / timeline & 24.8 & 12.5 & 13.4 & 16.3 & 14.0 \\
  & \# docs / timeline & 5.5 & 4.8 & 3.7 & 8.5 & 5.8 \\
\hline
\hline
  \multirow{10}{*}{\rotatebox{90}{Cross-lingual}} & \# documents & 60 & 60 & 60 & 60 & 180 \\
  & \# sentences & 917 & 891 & 861 & 926 & 2,678 \\
  & \# tokens & 21,208 & 20,898 & 21,116 & 21,257 & 63,271 \\
  \cline{2-7}
  & \# events mentions & 364 & 490 & 408 & 520 & 1,418 \\
  & \# event instance & 165 & 181 & 174 & 231 & 586 \\
  & \# target entities & 6 & 13 & 12 & 13 & 38 \\
  & \# timelines & 6 & 13 & 11 & 13 & 37 \\
  & \# events / timeline & 60.7 & 37.7 & 37.1 & 40.0 & 38.3 \\
  & \# event chains / timeline & 27.5 & 13.9 & 15.8 & 16.2 & 15.8 \\
  & \# docs / timeline & 11.5 & 10.0 & 8.2 & 17.6 & 12.1 \\
\hline
\end{tabular}
\caption{\label{tab:dataset} Counts extracted from the Multilingual and Cross-lingual gold datasets.}
\end{table*}

The English dataset released for the SemEval 2015 Timeline extraction task consists of 120 Wikinews\footnote{\url{http://en.wikinews.org}} articles containing 44 target entities. The Wikinews articles are focused mostly on four main topics, 30 documents per topic. A split of 30 documents and 6 target entities (each associated to a timeline) are provided as trial data, while the rest is left as evaluation set: 90 documents and 38 target entities (each associated to a timeline). Similarly, the Spanish dataset also contains 120 articles with 44 entities. The trial and test splits for this language are the same as in the English dataset. On the other hand, as the cross-lingual dataset arises from joining the English and Spanish datasets, it contains 240 articles containing same 44 target entities as in the English and Spanish datasets. In this case, the trial split includes 60 documents and 6 target entities while the test set contains the remaining 180 documents and 38 target entities.  For all the cases, the trial data contains one ORGANISATION target entity, one PERSON, and 4 PRODUCT entities. With respect to the evaluation set, 18 entities are ORGANISATION, 10 FINANCIAL, 7 PERSON, and 3 of the PRODUCT class. The four topics are the following: (i) Apple Inc. for the trial corpus; (ii) Airbus and Boeing; (iii) General Motors, Chrysler and Ford; and (iv) Stock Market.

Table~\ref{tab:dataset} provides some more details about the datasets. It should be noted that although there are 38 target entities, 37 were used for the evaluation because one timeline contained no events. Furthermore, although the three evaluation corpora are quite similar, the timelines created from the Stock Market corpus contain a higher average number of events with respect to those created from the other corpora. Additionally, it can also be seen that the Stock Market timelines contain events from a higher number of different documents. It should also be noticed that although the English and Spanish corpora are parallel translations the number of event instances and mentions in both cases are not exactly the same. This is due to the fact that some of the events from the English corpus cannot be expressed in Spanish with events that comply with the restrictions explained in Section \ref{sec:creation}. For example, in the sentence \textit{``Apple Computer would not sell music branded with an apple.''}, \textit{branded} can be included as an event mention in the \textit{Apple Computer} timeline because \textit{Apple Computer} is the \textit{Agent} of \textit{branded}. However, in the corresponding translated sentence \textit{``Apple Computer no vender\'ia m\'usica con una manzana por marca.''}, it is not possible to identify the \textit{Agent} of \textit{marca}, i.e. the translation of \textit{branded}.

\section{Evaluation Methodology}\label{sec:scorer}

The evaluation methodology proposed in SemEval 2015 was based on the evaluation metric used for TempEval3 \citep{UzZamanSemEval13}. The metric aims at capturing the temporal awareness of an annotation by checking the identification and categorization of temporal relations. In order to do this, \cite{UzZamanSemEval13} compare the graph formed by the relations given by a system ($Sys_{relation}$) and the graph of the reference (gold standard) annotations ($Ref_{relation}$). From these graphs, their closures ($Sys_{relation}^{+}$, $Ref_{relation}^{+}$) and reduced forms ($Sys_{relation}^{-}$, $Ref_{relation}^{-}$) are obtained. The reduced form is created by removing redundant relations (those that can be inferred from other relations) from the original graph. In this setting, Precision and Recall metrics are then calculated as follows:

$$ Precision = \frac{ | Sys_{relation}^{-} \cap Ref_{relation}^{+} | } { | Sys_{relation}^{-} | } $$
$$ Recall = \frac{ | Ref_{relation}^{-} \cap Sys_{relation}^{+} | } { | Ref_{relation}^{-} | } $$

Precision is calculated by counting the number of relations in the reduced system graph ($Sys_{relation}^{-}$) that can be found in the closure reference graph ($Ref_{relation}^{+}$) out of total number of relations in the reduced system graph ($Sys_{relation}^{-}$). Recall corresponds to the number of relations in the reduced reference graph ($Ref_{relation}^{-}$) that can be verified from the closure system graph ($Sys_{relation}^{+}$) out of the total number of relations in the reduced reference graph ($Ref_{relation}^{-}$). At the original SemEval 2015 task the following steps were proposed to transform the timelines into graphs of temporal relations:

\begin{enumerate}
\item Every time anchor is represented as a TIMEX3.
\item Each event is related to one TIMEX3 by means of the SIMULTANEOUS relation type.
\item If one event occurs before another one, a BEFORE relation type is created between both events.
\item If one event occurs at the same time as other event, a SIMULTANEOUS relation type links both events.
\end{enumerate}

\begin{figure*}[ht!]
  \centerline{
    \includegraphics[width=140mm]{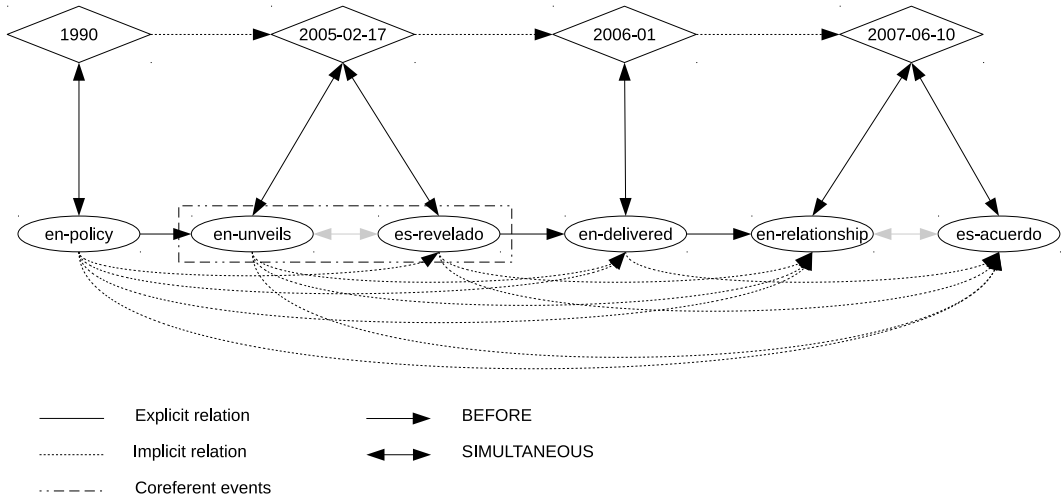}
    }
  \caption{\label{fig:SM15graph} Time graph produced by original SemEval 2015 evaluation. Grey lines represent redundant relations.}
\end{figure*}

Figure \ref{fig:SM15graph} shows the resulting graph after applying these four steps to the cross-lingual timeline in Figure \ref{fig:timeline}. The doted lines represent the implicit relations that will be part of the closure, while the grey lines represent the redundant relations absent in the reduced graph. For example, the SIMULTANEOUS relation between {\it en-unveils} and {\it es-revelado} can be inferred from the fact that both events are linked to the same TIMEX3 anchor via a SIMULTANEOUS relation.

Final scores are based on the micro-average of the individual $F_{1}$ scores for each timeline, namely, the scores are averaged over the events of the timelines of each corpus. The micro-averaged precision and recall values are also provided.

However, it is important to note that this evaluation method does not distinguish coreferent events, namely, mentions of the same event, from those that simply occur at the same time (simultaneous). In this sense, in Figure \ref{fig:SM15graph}, the same SIMULTANEOUS relation is used to connect two \emph{coreferent events} such as \emph{en-unveils} and \emph{es-revelado}, and two events {\it en-relationship} and {\it es-acuerdo}, that simply occur at the same time (e.g., they are not coreferent). Hence, while this methodology is sufficient to check the temporal ordering of events, it is not adequate for cross-lingual timeline extraction, because it is crucial to identify that two event mentions refer to the same event across languages. In order to address this issue, this paper extends the original evaluation method from the Timeline Extraction SemEval 2015 task and proposes two alternative scoring methods.

\subsection{Strict evaluation}\label{sec:strict-evaluation}

In the strict evaluation method a timeline must contain every mention of the events that can be found in the document set. Moreover, event mentions referring to the same event should be identified and distinguished from those that simply occur at the same time. With this aim in mind, the following changes are proposed:

\begin{itemize}
\item Coreferent events are not linked via the SIMULTANEOUS relation but by means of a new IDENTITY relation.
\item The IDENTITY relations are never removed from the reduced graphs. They are not redundant.
\end{itemize}

\begin{figure*}[ht!]
  \centerline{
    \includegraphics[width=140mm]{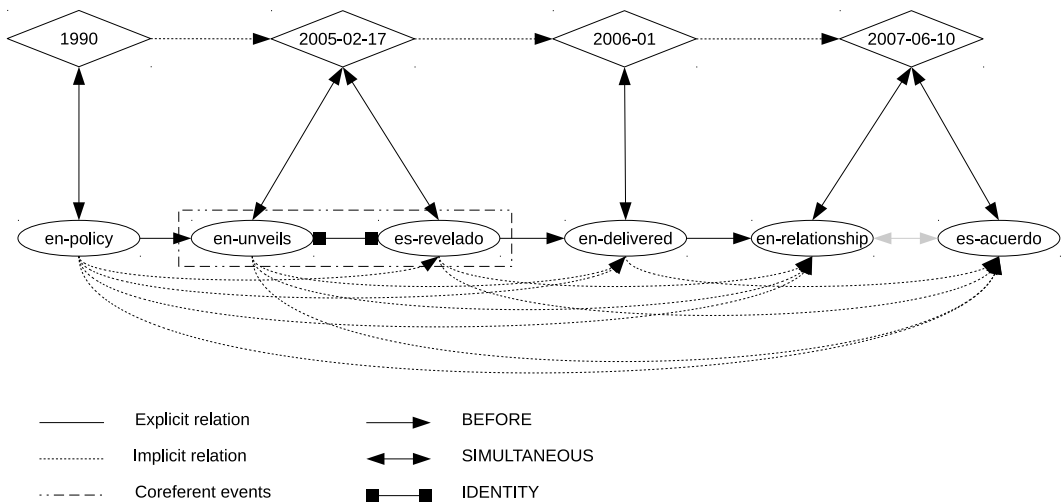}
    }
  \caption{\label{fig:strictgraph} Time graph produced by \textit{Strict evaluation}. Grey lines represent redundant relations.}
\end{figure*}

The strict temporal graph depicted in Figure \ref{fig:strictgraph} shows the graph obtained applying our new methodology. Whereas in the original graph in Figure \ref{fig:SM15graph} the coreferent events {\it en-unveils} and {\it es-revelado} are linked by a redundant SIMULTANEOUS relation, in Figure \ref{fig:strictgraph} a non-redundant IDENTITY relation links those two events.

Note that this method is more demanding in terms of precision because it adds the extra difficulty of distinguishing between IDENTITY and SIMULTANEOUS relations. Moreover, the set of temporal relations that must be captured is larger because the IDENTITY relations will not be removed when producing the reduced graphs. Thus, this also makes the task more demanding in terms of recall. That is why this evaluation method is named \textbf{strict evaluation}.

\subsection{Relaxed evaluation}\label{sec:relaxed-evaluation}

A second alternative stems from considering that, instead of using every event mention, a timeline could be composed of event types. Thus, coreferent events would be grouped as a single event by removing their temporal relations. The following changes are then performed with respect to the original SemEval 2015 evaluation:

\begin{itemize}
\item Every relation between coreferent events is removed.
\item All the SIMULTANEOUS relations between coreferent events and a TIMEX3 anchor are reduced to a single relation.
\end{itemize}

\begin{figure*}[ht!]

  \centerline{
    \includegraphics[width=140mm]{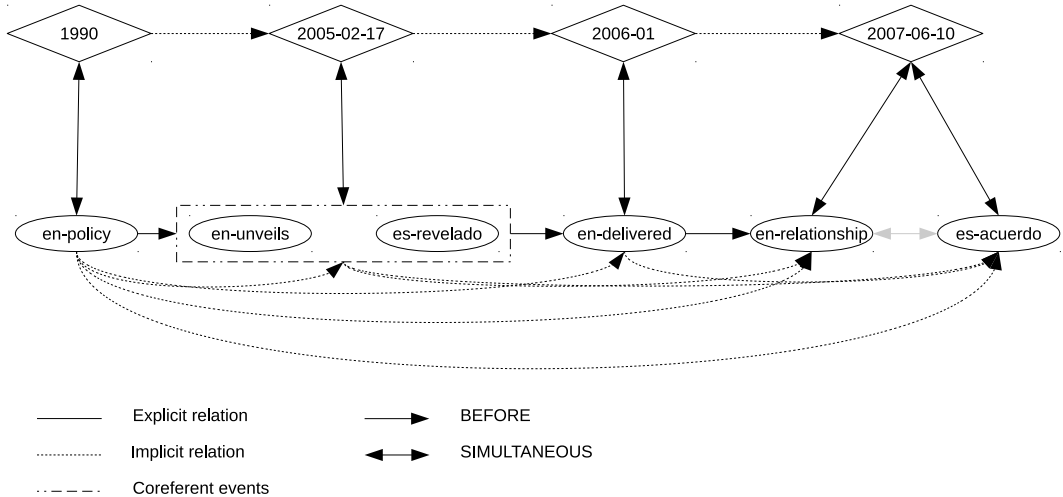}
    }
  \caption{\label{fig:relaxedgraph} Time graph produced by {\it Relaxed evaluation}. Grey lines represent redundant relations.}
\end{figure*}

These changes are explicitly shown by Figure \ref{fig:relaxedgraph}. It can be seen that there is no relation linking the {\it en-unveils} and {\it es-revelado} coreferent events. Furthermore, the SIMULTANEOUS relations that connected those event with their TIMEX3 have been reduced to one, namely, they are now linked to the event type (or to every mention of one specific event).

In this method the number of relations that must be captured is smaller because detecting just one of the coreferent event mentions shall be enough. Thus, this evaluation is more \textbf{relaxed} in terms of recall. However, it is still required to properly detect coreferent events, otherwise they will be evaluated as different instances, consequently harming the precision.

\section{Automatic Cross-lingual TimeLine extraction}\label{sec:system}

This section presents our approach for timeline extraction, including both multilingual and cross-lingual systems. Given a set of documents and a target entity, a three step process is applied. First, the mentions of the target entity are identified. Second, the events in which the target entity is involved are selected. Finally, those events are anchored to their respective normalized time expressions. Once this process is completed, the events are sorted and the timeline built.

In the following we describe the three different systems for Timeline extraction applied to the tasks previously described. Section \ref{sec:bte} introduces the baseline (BTE) system. BTE performs timeline extraction by combining the output of a NLP pipeline for both English and Spanish. The baseline system is then improved in section \ref{sec:dlt} by applying the algorithm presented in \cite{LaparraACL2015} to perform document level time-anchoring (DLT). While both BTE and DLT can be used for multilingual timeline extraction, their performance in the cross-lingual setting is not as good as in the English and Multilingual tasks. Thus, in section \ref{sec:cross-ev} we propose a new approach to obtain interoperable annotations across languages from the same NLP pipelines used for BTE in section \ref{sec:bte}. We can then use this approach to identify coreferent event mentions across languages which is crucial to build cross-lingual timelines.

\subsection{BTE: Baseline TimeLine Extraction}\label{sec:bte}

Detecting mentions of events, entities and time expressions in text requires the combination of various NLP tools. We apply the NewsReader NLP pipelines \citep{VossenKBS2016} that includes, both for English and Spanish, Named Entity Recognition (NER) and Disambiguation (NED), Coreference Resolution (CR), Semantic Role Labelling (SRL), Time Expressions Identification (TEI) and Normalization (TEN), and Temporal Relation Extraction (TRE). Table \ref{tab:nlp-pipelines} lists the specific tools used for English and Spanish.

\begin{table*}[ht!]
{\footnotesize
  \begin{center}
    \begin{tabular}{l|l|l} \hline
      & \multicolumn{1}{c|}{English} & \multicolumn{1}{c}{Spanish} \\ \hline \hline
      NER & \multicolumn{2}{c}{\cite{AgerriNER}} \\ \hline
      NED & \multicolumn{2}{c}{\cite{spotlight}} \\ \hline
      CR & \multicolumn{2}{c}{\cite{agerri_ixa_2014}} \\ \hline
      SRL & \multicolumn{2}{c}{\cite{Bjorkelund:2009}} \\\hline
      TEI & \cite{mirzaEVENTIevalita} & \cite{Strotgen:2013} \\
      \hline
      TEN & \cite{mirzaEVENTIevalita} & \cite{Strotgen:2013} \\
      \hline
      TRE & \cite{MirzaEACL14} & \cite{llorens2010tipsem} \\
      \hline
    \end{tabular}
    \caption{English and Spanish NLP tools.}
    \label{tab:nlp-pipelines}
  \end{center}}
\end{table*}

The extraction of target entities, events and time anchors is performed as follows:

\textbf{(1) Target entity identification}: The target entities are identified by the NER and NED modules. As the surface form of the candidate entities can vary greatly, we use the redirect links contained in DBpedia to extend the search of the events involving those target entities. For example, if the target entity is \emph{Toyota} the system would also include events involving the entities \emph{Toyota Motor Company} or \emph{Toyota Motor Corp.} In addition, as the NED does not always provide a link to DBpedia, we also check if the wordform of the head of the event argument matches with the head of the target entity.

\textbf{(2) Event selection}: We use the output of the SRL module to extract the events that occur in a document. Given a target entity, we combine the output of the NER, NED, CR and SRL to obtain those events that have the target entity as filler of their ARG$0$ or ARG$1$. We also set some constraints to select certain events according to the specification of the SemEval task. Specifically, we only return those events that are not within the scope of a negation and that are not accompanied by modal verbs (except \emph{will}).

\textbf{(3) Time-anchoring}: The time-anchors are identified using the TRE and SRL output. From the TRE, we extract as time-anchors
those relations between events and time expressions identified as SIMULTANEOUS. From the SRL those ARG-TMP related to time expressions. In both cases we use the time expression returned by the TEI module. The tests performed on the trial data showed that the best choice for time-anchoring results from combining both options. For each time-anchor we normalize the time expression using the annotations provided by the TEN module.

\subsection{DLT: Document Level Time-anchoring}\label{sec:dlt}

\begin{figure*}[ht!]
  \centering
  \includegraphics[width=120mm]{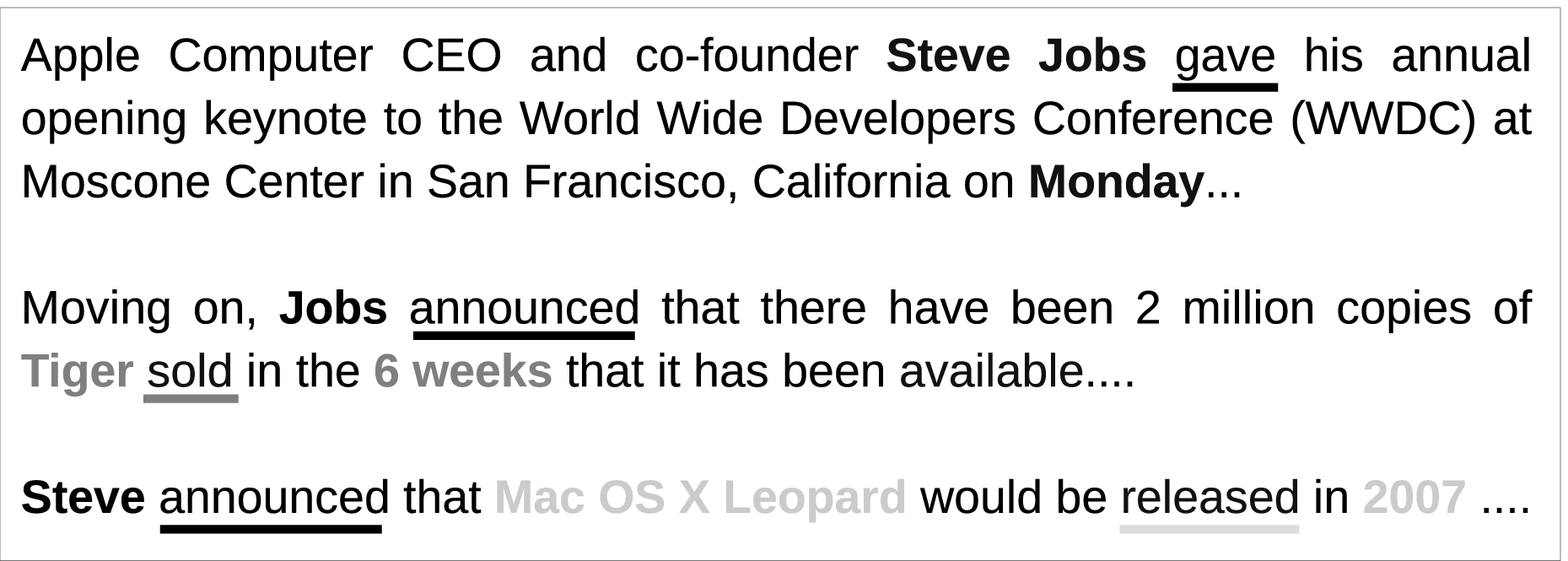}
  \caption{\label{fig:dlt-example} Example of time-anchoring at document level.}
\end{figure*}

The explicit time anchors provided by the NLP tools presented in previous section \ref{sec:bte} do not cover the full set of events involving a particular entity. In other words, most events do not have an explicit time anchor and therefore are not captured as part of the timeline of that entity. This means that we have to be able to also recover those time-anchors that are implicitly conveyed in the text.

In \cite{LaparraACL2015} we devised a simple strategy to capture implicit time-anchors while maintaining the coherence of the temporal information in the document. The rationale behind the algorithm shown in Algorithm \ref{alg:ImplicitTimeLines} is that, by default, the events of a specific entity that appear in a document tend to occur at the same time as previous events involving the same entity (unless explicitly stated). For example, in Figure \ref{fig:dlt-example} every event related to \textit{Steve Jobs}, such as \textit{gave} and \textit{announced}, are anchored to the same time expression (\textit{Monday}) even though it is only explicitly conveyed for the first event \textit{gave}. This example also illustrates the fact that for those other events that occur at different times, their time-anchor is also explicit, as it can be seen for the \textit{Tiger} and \textit{Mac OS X Leopard} entities.

The application of Algorithm \ref{alg:ImplicitTimeLines} starts taking as input the annotation obtained by the NLP described in Section \ref{sec:bte}. For each entity a list of events ($eventList$) is created sorted by appearing order. Next, for each event in the list the DLT system checks whether that event already has assigned a time-anchor ($eAnchor$). If that is the case, that time-anchor is included in the list of default time-anchors ($defaultAnchor$) for any subsequent events of the entity in the same verb tense ($eTense$). If the event does not yet have an explicit time-anchor assigned, but the system has found a time-anchor for a previous event in the same tense ($defaultAnchor[eTense]$), this time-anchor is also assigned to the current event ($eAnchor$). If none of the previous conditions hold, then the algorithm anchors the event to the \textbf{Document Creation Time} (DCT) attribute and sets this time-expression as the default time-anchor for any subsequent events in the same verbal tense.

\begin{algorithm}
\caption{\label{alg:ImplicitTimeLines}Implicit Time-anchoring}
\begin{algorithmic}[1]
\STATE $eventList$ = sorted list of events of an entity
\FOR{$event$ in $eventList$}
\STATE $eAnchor$ = time anchor of $event$
\STATE $eTense$ = verb tense of $event$
    \IF{$eAnchor$ not $NULL$}
	\STATE $defaultAnchor[eTense]$ = $eAnchor$
    \ELSIF{$defaultAnchor[eTense]$ not $NULL$}
	\STATE $eAnchor$ = $defaultAnchor[eTense]$
   \ELSE
	\STATE $eAnchor$ = DCT
	\STATE $defaultAnchor[eTense]$ = DCT
    \ENDIF
\ENDFOR
\end{algorithmic}
\end{algorithm}

The \textbf{DLT} system build the timeline by ordering the events according to the explicit and implicit time-anchors. Note that Algorithm \ref{alg:ImplicitTimeLines} strongly depends on the tense of the mentions of events appearing in the document. As this information can be only recovered from verbal predicates, this strategy cannot be applied to events conveyed by nominal predicates. Consequently, for these cases just explicit time-anchors are taken into account.

\subsection{CLE: Cross-Lingual Event coreference}\label{sec:cross-ev}

As it has been already mentioned, cross-lingual timeline extraction crucially depends on being able to identify those events that are coreferent across languages (not only across documents). In order to address this issue, we propose a language independent knowledge representation for cross-lingual semantic inter-operability at three different annotation levels.

First, we used interconnected links in the DBpedia entries to perform cross-lingual Named Entity Disambiguation (NED). The NED module used in the NLP pipeline for BTE provides links to the English and Spanish versions of the DBpedia. Thus, a mention of \textit{New York} in English should link as external reference to the the identifier \url{http://dbpedia.org/page/New_York}. Similarly, a mention of \textit{Nueva York} in Spanish should produce as external reference the identifier \url{http://es.dbpedia.org/page/Nueva_York}. As both identifiers are connected within the DBpedia, we can just infer that those two pointers refer to the same target entity regardless of the language in which the mentions of that entity are expressed.

Second, we obtain inter-operability across languages and Semantic Role Labeling annotations by means of the PredicateMatrix \citep{LopezdeLacalleLREC16, LopezdeLacalleLRE16}. The event representation provided by our SRL systems are based on PropBank, for English, and AnCora \citep{Taule08}, for Spanish. The PredicateMatrix gathers knowledge bases that contain predicate and semantic role information in different languages, including links between PropBank and AnCora. Using these mappings, we can establish, for example, that the role {\it arg0} of the Spanish predicate {\it vender.1} is aligned to the role {\it A0} of the PropBank predicate {\it sell.01}.

Finally, the TEN modules normalize time expressions following the ISO 24617-1 standard \citep{pustejovsky2010isotimeml}. For example, if temporal expressions such as {\it next Monday}, {\it tomorrow}, and {\it yesterday} in English or {\it ayer} and {\it el pr\'oximo lunes} in Spanish are referring to the same exact date (let's say {\it November 16th, 2015}), then they will be normalized to the same TIMEX3 value corresponding to {\it 2015-11-16}.

We can include these three levels of cross-lingual information to extend the multilingual system DLT presented in the previous section. When extracting the cross-lingual timeline for a given target entity, expressed as $e_{E}$ and $e_{S}$ in English and Spanish respectively, the system establishes that the English event $p_{E}$ and the Spanish event $p_{S}$ are coreferent if the following conditions are satisfied:

\begin{enumerate}
\item $e_{E}$ and $e_{S}$ are connected by DBpedia links to the same entity.
\item $e_{E}$ plays the role $r_{E}$ of $p_{E}$, $e_{S}$ plays the role $r_{S}$ of $p_{S}$, and $r_{E}$ and $r_{S}$ are linked by a mapping in the PredicateMatrix.
\item $p_{E}$ is anchored to a TIMEX3 $t_{E}$, $p_{S}$ is anchored to a TIMEX3 $t_{S}$ and $t_{E}$ and  $t_{S}$ are normalized to the same ISO 24617-1.
\end{enumerate}

The \textbf{CLE} system uses the same strategy as DLT to build timelines with the difference that cross-lingual coreferent events are identified.

\section{Experimental Results}\label{sec:evaluation}

In this section we present a set of experiments in order to evaluate the three timeline extraction systems presented in the previous section: (i) the \textbf{BTE} baseline system based on the analysis given by a pipeline of NLP tools; (ii) the \textbf{DLT} algorithm that aims at capturing implicit time-anchoring at document level; and (iii) the \textbf{CLE} system to address cross-lingual event co-reference. The evaluations are undertaken for the original English SemEval 2015 task as well as for the Multilingual and Cross-Lingual Timeline Extraction tasks proposed in section \ref{sec:mult-crossl-timel}. Every result is evaluated using the original SemEval 2015 metric as well as the \emph{strict} and \emph{relaxed} metrics introduced in section \ref{sec:scorer}.

\subsection{Multilingual evaluation}\label{sec:mult-eval}

In this setting we evaluate both BTE and DLT systems on the Track A (main track) of the TimeLine Extraction task at SemEval 2015 and on the Multilingual task described in section \ref{sec:mult-crossl-timel}. Track A at SemEval 2015 had just two participant teams, namely, \textbf{WHUNLP} and \textbf{SPINOZAVU}, which submitted three runs in total. Their scores in terms of Precision (P), Recall (R) and F1 are presented in Table \ref{tab:resultsSE}. We also present in italics additional results of both systems obtained after the official evaluation task \citep{caselli2015}. The best run was obtained by the corrected version of \textbf{WHUNLP\_1} with an F1 of 7.85\%. The low figures obtained show the difficulty of the task.

\begin{table}[h]
\centering
\begin{tabular}{lrrr}
 System & P & R & F1\\
\hline
SPINOZAVU-RUN-1 &  7.95 & 1.96 & 3.15 \\
SPINOZAVU-RUN-2 &  8.16 & 0.56 & 1.05 \\
WHUNLP\_1 	& 14.10 & 4.90  & 7.28 \\
\hline
\hline
\emph{OC\_SPINOZA\_VU} & - & - & 7.12 \\
\emph{WHUNLP\_1} & 14.59 & 5.37 &  7.85 \\
\hline
\hline
\textbf{BTE} & \textbf{24.56} & 4.35 & 7.39 \\
\textbf{DLT} & 21.00 & \textbf{11.01} & \textbf{14.45} \\
\end{tabular}
\caption{Results on the SemEval-2015 task}
\label{tab:resultsSE}
\end{table}

Table \ref{tab:resultsSE} also contains the results obtained by our systems. The results obtained by our baseline system, \textbf{BTE}, are similar to those obtained by \textbf{WHUNLP\_1}. However, the results of the implicit time-anchoring approach (\textbf{DLT}) clearly outperforms our baseline and every other previous result in this task. This result would imply that a full time-anchoring annotation requires that a temporal analysis be carried out at document level. As expected, Table \ref{tab:resultsSE} also shows that the improvement of DLT over BTE is much more significant in terms of Recall.
\begin{table}[h]
\centering
  \begin{tabular}{ll rrr || rrr}
    & & \multicolumn{3}{c||}{English} &  \multicolumn{3}{c}{Spanish}\\
    \hline
    Scorer & System & P & R & F1 & P & R & F1 \\
    \hline
    \multirow{3}{*}{SemEval-2015} & \textbf{BTE} & 24.56 & 4.35 & 7.39 & 12.07 & 4.25 & 6.29 \\
    & \textbf{DLT} & 21.00 & 11.01 & 14.45 & 12.77 & 8.60 & 10.28 \\
    \hline
    \multirow{3}{*}{strict-evaluation} & \textbf{BTE} & 24.56 & 3.62 & 6.32 & 12.07 & 3.60 & 5.55 \\
    & \textbf{DLT} & 21.00 & 9.18 & 12.77 & 12.77 & 7.29 & 9.28 \\
    \hline
    \multirow{3}{*}{relaxed-evaluation} & \textbf{BTE} & 24.12 & 5.32 & 8.71 & 11.55 & 5.18 & 7.15 \\
    & \textbf{DLT} & 19.39 & 12.95 & 15.53 & 11.47 & 9.72 & 10.52 \\
    \hline
  \end{tabular}
\caption{Results on the multilingual task.}
\label{tab:resultsML}
\end{table}

Table \ref{tab:resultsML} provides the results obtained by \textbf{BTE} and \textbf{DLT} in the Multilingual Timeline extraction setting using also the \emph{strict} and \emph{relaxed} evaluation metrics described in Section~\ref{sec:scorer}. Predictably, the strict evaluation is the most demanding, specially in terms of Recall. With respect to the results obtained using the relaxed scorer, precision is lower whereas recall is higher with respect to the other two metrics. Furthermore, \textbf{DLT} outperforms \textbf{BTE} whatever the language and the evaluation methodology. It is also remarkable that the results obtained for English are always better than the results for Spanish. This can be explained by the differences in the performances of the English and Spanish NLP modules.

\subsection{Cross-lingual evaluation}\label{sec:cross-ling-eval}

The dataset for cross-lingual timelines contains 180 documents (see Section \ref{sec:annotation}), of which half are Spanish translations of the other half written in English. This fact allows us to set different experiments by varying the percentage of documents written in each language that are provided as input. Three different experiments were performed in order to evaluate our systems on the Cross-lingual Timeline extraction task.

\paragraph{Full data} For the first experiment, we use as input the full collection (180 documents) independently of the language. As shown by Table \ref{tab:resultsCL}, the results using the SemEval 2015 scoring method, as it was the case in the multilingual setting, the \textbf{DLT} system almost doubles the score of the baseline system \textbf{BTE}. Furthermore, \textbf{DLT} and \textbf{CLE} obtain exactly the same results because co-referent events are not taken into account. However, the \emph{strict} and \emph{relaxed} scoring methods proposed in this work make it possible to distinguish between the performances of the two systems. Not surprisingly, the scoring by strict evaluation continues to be the lowest. Overall, \textbf{CLE} outperforms \textbf{DLT} being only in terms of precision (relaxed evaluation) or in both precision and recall (strict evaluation).

\begin{table}[h]
\centering
  \begin{tabular}{ll | rrr}
    Scorer & System & P & R & F1 \\
    \hline
    \multirow{2}{*}{SemEval-2015} & \textbf{BTE} & 13.98 & 4.68 & 7.02 \\
    & \textbf{DLT} & 14.96 & 10.74 & 12.50 \\
    & \textbf{CLE} & 14.96 & 10.74 & 12.50 \\
    \hline
    \multirow{2}{*}{strict-evaluation} & \textbf{BTE} & 13.98 & 3.12 & 5.10 \\
    & \textbf{DLT} & 14.96 & 7.14 & 9.67 \\
    & \textbf{CLE} & 16.59 & 8.47 & 11.22 \\
    \hline
    \multirow{2}{*}{relaxed-evaluation} & \textbf{BTE} & 10.13 & 8.16 & 9.04 \\
    & \textbf{DLT} & 9.75 & 17.70 & 12.57 \\
    & \textbf{CLE} & 10.97 & 17.70 & 13.55 \\
    \hline
  \end{tabular}
\caption{Results on the cross-lingual task}
\label{tab:resultsCL}
\end{table}

\begin{figure*}[ht!]
  \centerline{
    \includegraphics[width=130mm,height=150mm]{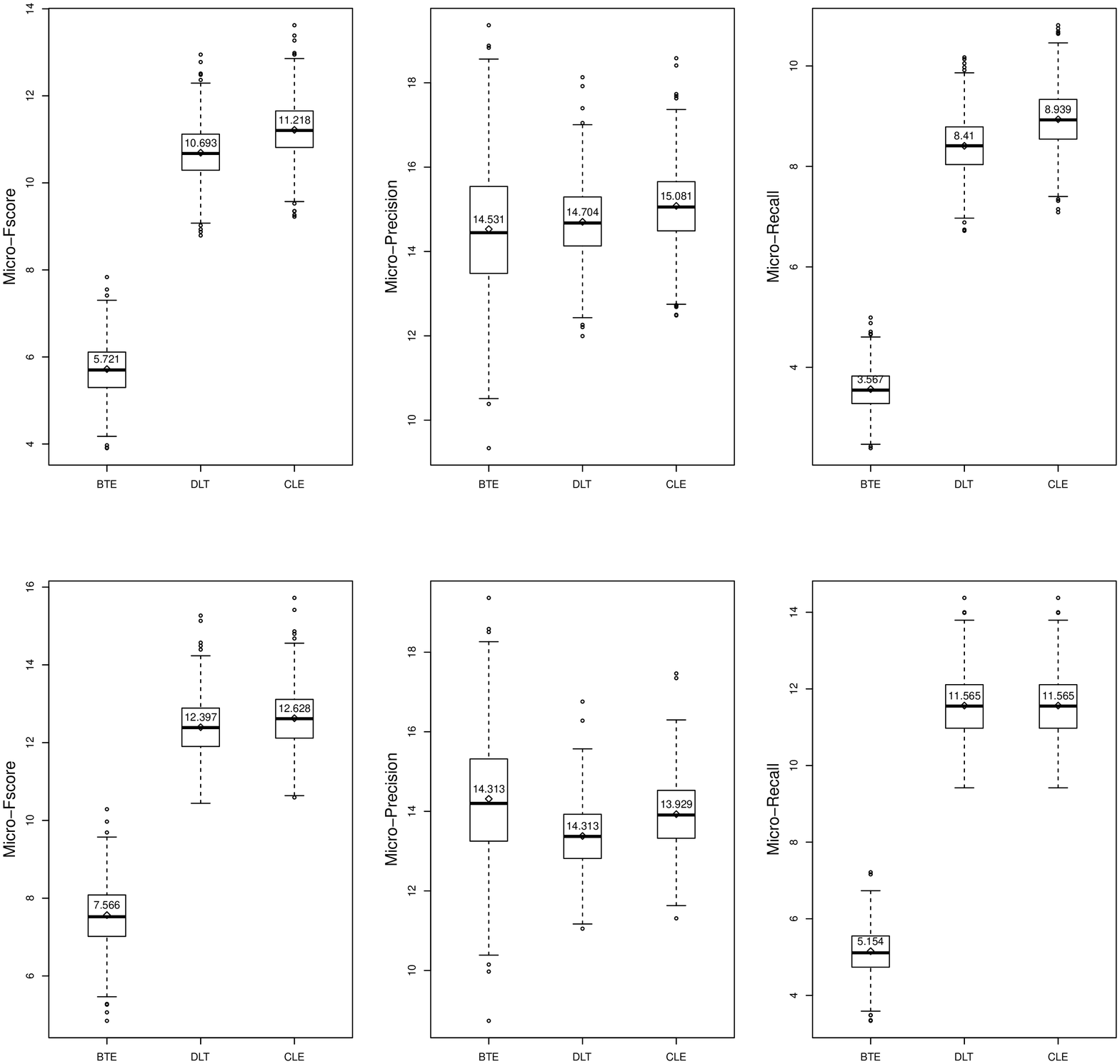}
    }
  \caption{\label{fig:results50} Evaluation 50-50. The top row results are calculated using the \emph{strict} metric whereas the results at the bottom row refer to the \emph{relaxed} evaluation method.}
\end{figure*}

\paragraph{50-50 split} As we believe that the availability of a set of parallel documents as input is not the most realistic scenario, we design another setting by choosing at random 50\% of the documents in each language, namely, 45 documents for English and 45 for Spanish respectively.
The resulting input set would contain 90 non-parallel documents in two languages without the mentions of the events that belong to documents not included in the final collection of 90 documents. Furthermore, we automatically generate not just one but $1,000$ different 50-50 input sets of 90 documents at random, namely, each of the thousand sets contain 45 documents in each language. The box-plots in Figure \ref{fig:results50} show the results obtained by our systems in this experiment applying the strict and relaxed evaluation methodologies to the one thousand evaluation sets.

Following the trend of previous results, both \textbf{DLT} and \textbf{CLE} outperform the baseline system with \textbf{CLE} obtaining the best overall performance. The F1 score differences between \textbf{DLT} and \textbf{CLE} using both evaluation methods are significant with $p < 0.001$.\footnote{We have used the paired $t$-test to compare the $F1$ obtained by the systems.} In any case, the results show that performance between \textbf{DLT} and \textbf{CLE} has reduced with regard to the results obtained in the previous experiment reported by Table \ref{tab:resultsCL}. Our hypothesis is that as the set of input documents in this experiment has been halved, the number of coreferent mentions in the gold-standard is much lower, which means that the advantage of CLE over DLT is not that meaningful. The most remarkable variation can be observed in the Recall values obtained using the relaxed evaluation. This is not that strange if we consider that in the relaxed evaluation detecting only one mention of an event is enough.

\begin{figure*}[ht!]
  \centerline{
    \includegraphics[width=140mm]{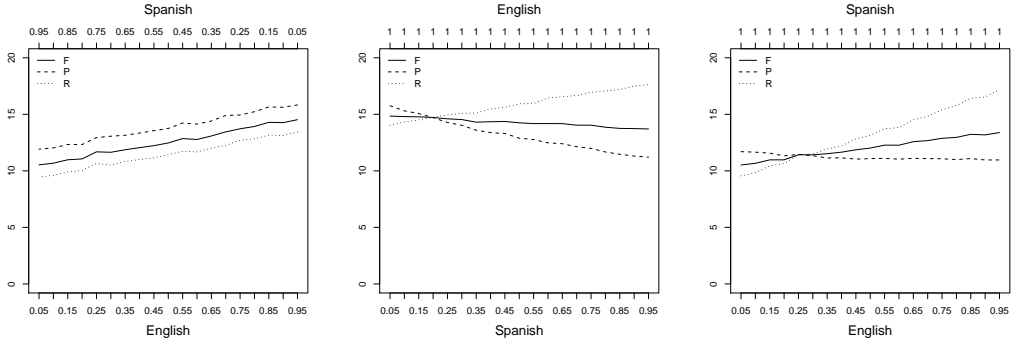}
    }
  \caption{\label{fig:resultsVar} Varying the number of input documents per language. The $y$ axis en each box represents the percantage of documents used for each language.}
\end{figure*}

\paragraph{Varying input per language} This last experiment was designed to study how varying the number of documents per language affects the performance in the cross lingual setting. The line charts in Figure \ref{fig:resultsVar} show the results obtained varying the percentage of the documents being used.

On the left-hand side plot we show the results of experiments using a range of 5\% to 95\% documents for both languages (Spanish on top, English at the bottom). Now, for each point in the range we randomly generate 30 input sets. For example, at the 10\% Spanish and 90\% English 30 different configurations are randomly generated each of which would contain 81 English documents and 9 Spanish documents (90 documents in total).

In the experiments reported by the central and right-hand size plots, we use the above method to generate 30 input sets for each point in the range, but with two important differences. Firstly, every document in one language is alternatively used (English in the central plot and Spanish on the right-hand side) and we increase the number of documents in the other language from 5\% up to 95\% (when the 180 documents are used). Secondly, in these two cases parallel documents are allowed.

For all three cases each point represents the {\it arithmetic mean} of the output given for the 30 different input document sets generated without replacement. The evaluation method used is \emph{relaxed} due to the fact that we start with the full set of possible events. Thus, varying or increasing the number of documents in the other language does not in fact increase the number of events, just (possibly) the number of event mentions. Therefore, the \emph{relaxed} method allows us to focus on studying whether adding parallel documents in other language improves the overall F1 score, paying particular attention to the Recall.

The results illustrate that the CLE F1 score keeps degrading as we include Spanish documents into the fold. This is somewhat explained by CLE results obtained in Table \ref{tab:resultsML} where the performance of the Spanish system is much worse that its English counterpart.

\section{Error Analysis}\label{sec:error}

As shown in Section \ref{sec:bte}, our baseline approach for timeline extraction (on which \textbf{DLT} and \textbf{CLE} build) is based on the output of a set of NLP modules. Now, although they are state-of-the-art tools on standard evaluation data, they still produce cascading errors, most notably when applied to out-of-domain data (see Table 8 at \cite{VossenKBS2016}). The aim of this section is to identify the main source of errors.

\begin{table*}[ht!]
  \begin{center}
  \begin{tabular}{l|r|r|r}
    & English & Spanish & Cross-lingual \\
    \hline
    full (BTE) & 6.04 & 8.43 & 8.20 \\
    full (DLT) & 19.55 & 17.70  & 19.83 \\
    \hline
    SRL & 72.06 & 56.29 & 63.04 \\
    SRL+NER+NED & 22.01 & 17.67 & 20.97 \\
    SRL+NER+NED+CR & 23.95 & 17.67 & 22.25 \\
    \hline
    SRL+TEI+TEN+TRE (BTE) & 13.72 & 21.50 & 19.47 \\
    SRL+TEI+TEN+TRE (DLT) & 46.16 & 53.21 & 50.43 \\
    \hline
  \end{tabular}
  \caption{Percentage of events captured by the pipelines. The \textit{full} rows correspond to SRL+NER+NED+CR+TEI+TEN+TRE.}
  \label{tab:error-event}
  \end{center}
\end{table*}

In a first experiment we study the capability of our system for extracting those events that participate in the timelines, regardless of time ordering. The first two rows in Table \ref{tab:error-event} show that the \textbf{DLT} system is able to extract way more events than the \textbf{BTE} baseline system, however in both cases the percentage of events captured is still low. To study the causes of these figures we have repeated the same experiment with partial combinations of the NLP modules. As explained in Section \ref{sec:bte}, we use a SRL system to detect event mentions. Table \ref{tab:error-event} shows that for English the SRL module detects more events than for Spanish (72.06\% vs 56.29\%). This is largely due to the Spanish SRL not dealing correctly with verbal nominalizations.

In order to extract only those events that are linked to the target entity, we use the combined output of the SRL, NER, NED and CR tools (see Section \ref{sec:bte}). Table \ref{tab:error-event} shows that this is a very difficult step and that the percentage of events identified is rather low. Detecting and linking every mention of an entity is a very difficult task, specially in the case of pronouns. As it can be seen, the coreference module helps although not as much as it would have been expected.

The final two rows of Table \ref{tab:error-event} report on the results obtained when only events with a time anchor are included in a timeline. The number of events linked to a explicit time-anchor by our BTE baseline system is very low whereas looking at the implicit anchors in the DLT system helps to substantially improve the results. Notice that in this case the figures are higher for Spanish (21.50\% and 53.21\%) than for English (13.72\% and 46.16\%). This means that time modules for Spanish try to anchor more events that the English modules.

In a second experiment we study the quality of the time anchoring. Table \ref{tab:error-anchor} shows the accuracy of the time-anchors for the events that we know have been correctly identified. It makes sense that the accuracy of \textbf{DLT} be much lower than just taking into account explicit time-anchors as \textbf{BTE} does. However, it should be noted that number of events extracted by the DLT system is much higher than BTE (as per Table \ref{tab:error-event}), which means that accuracy for DLT is calculated over a much larger number of correctly identified event mentions. As can be seen, the English systems perform better than the Spanish systems. As explained above, the Spanish modules try to time-anchor more events and this fact can explain that they obtain a lower accuracy.

\begin{table*}[ht!]
  \begin{center}
    \begin{tabular}{l|r|r|r}
      & English & Spanish & Cross-lingual \\
      \hline
      BTE & 69.49 & 50.70 & 68.70 \\
      DLT & 51.31 & 46.98 & 62.59 \\
      \hline
    \end{tabular}
    \caption{Accuracy of the time-anchoring for extracted events.}
    \label{tab:error-anchor}
  \end{center}
\end{table*}


\section{Concluding Remarks}\label{sec:concluding-remarks}

In this work we present a system to perform Multilingual and Cross-lingual Timeline Extraction (or Cross-document event ordering). In doing so, this paper presents a number of novel contributions.

Firstly, the original Cross-document event ordering task defined for SemEval 2015 (main Track A) has been extended to present two novel tasks for two languages (English and Spanish) on Multilingual and Cross-lingual timeline extraction respectively. The annotated datasets for trial and evaluation are publicly available.

Secondly, two new evaluation metrics improve the evaluation methodology of the SemEval 2015 task in two ways: (i) A new \emph{strict} metric allows to evaluate timelines containing coreferent event mentions across both documents and languages; and (ii) a \emph{relaxed} evaluation metric where event types (instead of mentions) can be considered, somewhat diminishing the importance of recall when evaluating the timelines.

Thirdly, three deterministic Timeline extraction systems have been developed to address the three tasks. In fact, we have empirically demonstrated that addressing implicit time-anchors at document level (DLT system) crucially improves the performance in the three tasks, clearly outperforming previously presented systems in the (main) Track A of the original Timeline Extraction task at SemEval 2015. Furthermore, we have shown how to effectively use cross-lingual resources such as the PredicateMatrix and DBpedia along with time normalization to improve the performance of the DLT system in the most realistic setting of building cross-lingual timelines without parallel data as input (see Figure \ref{fig:results50}).

Finally, we have analyzed the cascading errors produced by the NLP pipeline used to identify the entities, events and time-anchors. The results allow to conclude that the most difficult obstacles reside in detecting and resolving every mention of entities related to the relevant mention events and the identification of time-anchors when they are not explicitly conveyed. These two aspects shall point out future work towards improving timeline extraction.

\section*{Acknowledgments}

This work has been supported by the European projects QTLeap (EC-FP7-610516) and NewsReader (EC-FP7-316404) and by the Spanish Ministry for Science and Innovation (MICINN), SKATER (TIN2012-38584-C06-01) and TUNER (TIN2015-65308-C5-1-R).

\end{document}